\documentclass[10pt]{article}
\usepackage[a4paper,margin=2cm,headheight=14pt]{geometry}
\usepackage[T1]{fontenc}
\usepackage{lmodern,microtype}
\usepackage{amsmath,amssymb}
\usepackage{graphicx,float,booktabs,tabularx,array}
\usepackage[labelfont=bf,font=small]{caption}
\usepackage{enumitem}
\usepackage[numbers,sort&compress]{natbib}
\usepackage{xcolor}
\usepackage[colorlinks=true,linkcolor=blue!50!black,citecolor=blue!50!black,urlcolor=blue!50!black]{hyperref}
\usepackage{fancyhdr}
\newcolumntype{L}{>{\raggedright\arraybackslash}X}
\newcolumntype{C}{>{\centering\arraybackslash}X}
\title{\bfseries Ensemble Classification and Cancer-Specific Gene Signatures from Quality-Aware Hybrid Gene Selection in TCGA RNA-Seq Data}
\author{Tareque Mohmud Chowdhury\textsuperscript{1,*} \and Abu Raihan Mostofa Kamal\textsuperscript{1}\\[6pt]
\small \textsuperscript{1}Department of Computer Science and Engineering, Islamic University of Technology,\\ \small Gazipur 1704, Bangladesh\\[2pt]
\small \textsuperscript{*}Correspondence: \href{mailto:tareque@iut-dhaka.edu}{tareque@iut-dhaka.edu}}
\date{}

\begin{document}
\maketitle
\thispagestyle{fancy}

\begin{abstract}
\noindent Gene-expression classifiers of cancer type are usually developed and validated within one cohort using random splits. It is then unclear whether complex models improve on a regularised linear model, and how much accuracy depends on the same tissue source sites appearing on both sides of the split. Using the panels of a quality-aware hybrid Boruta selector from our previous study, we evaluated ensembles, robustness to unseen sites, external transfer and tumour versus adjacent-normal gene signatures. Using 10,469 TCGA tumours of 33 types, we compared five learners and three ensembles under decision rules fixed in advance, with gene selection nested in every training fold. We reran the selection in cross-validation grouped by tissue source site under five site-to-fold assignments, transferred a breast-cancer subtype panel to 3273 tumours of the independent SCAN-B cohort, and built five-gene stage~I tumour versus adjacent-normal panels for nine cancer types. Logistic regression (LR) was the most accurate and best-calibrated of the five single learners (balanced accuracy 0.943 on the locked test set). Stacking added 0.0046 with 1000 genes, just below the pre-declared margin of 0.005, but improved on LR in every fold with a minimal 100-gene panel. Site-grouped validation lowered balanced accuracy by 0.010--0.027 across five site-to-fold assignments, mostly through colon and rectal cancer (0.003--0.008 without them and LAML). On SCAN-B, the subtype panel reached 0.855, comparable to the baseline panel and close to the 50 PAM50 genes (0.876). The five-gene panels reached ROC AUC $\geq$ 0.99 in held-out patients of the seven testable types and also detected later-stage tumours. With 1000 genes, no ensemble improved on a well-calibrated linear model by the pre-declared margin. Random splits overstate accuracy where site and label are entangled, chiefly in colorectal cancer. Site-grouped and external validation should be routine checks for TCGA-derived classifiers.

\end{abstract}

\section{Introduction}

RNA sequencing of tumour samples captures the transcriptional state of a tumour across tens of thousands of genes~\cite{wang2009}. The Cancer Genome Atlas (TCGA) has made such profiles available for more than 10,000 tumours of 33 cancer types~\cite{weinstein2013,grossman2016}, and its pan-cancer analyses showed that the cell of origin dominates the molecular classification of tumours~\cite{hoadley2018}. Machine-learning classifiers trained on these data reach high accuracy in separating cancer types~\cite{li2017,mostavi2020,mohammed2021,khorshed2020}, and recent reviews document a rapid growth of such methods~\cite{alharbi2023,stancl2023}. Classification of cancer type matters clinically for tumours of unknown primary, where molecular classifiers may help to guide treatment~\cite{moon2023}, and for molecular confirmation of a histological diagnosis.

Earlier work relevant to this study falls into three strands. First, gene selection methods for expression data are grouped into filters, which score genes one at a time; wrappers, which search gene subsets with a classifier; and embedded methods, which select genes while the model is fitted~\cite{saeys2007,ang2016}. Hybrid schemes chain a fast filter with a wrapper, so that the wrapper remains tractable on tens of thousands of genes~\cite{ang2016}; proposing such hybrids has been the most frequent research direction in recent gene selection studies~\cite{alhenawi2022,almazrua2022}. Among random-forest wrappers, Boruta~\cite{kursa2010} was one of the best-performing methods in a comparison on omics data~\cite{degenhardt2019}. Whatever the method, selection must be repeated inside every training fold, or accuracy is overestimated~\cite{ambroise2002,varma2006}. Second, many classifiers of cancer type have been trained on TCGA. Li et al.\ combined a genetic algorithm with $k$-nearest neighbours for 31 tumour types and classified more than 90\% of a random 25\% test split correctly~\cite{li2017}. Convolutional networks reached 93.9--95.0\% accuracy for 33 types and normal tissue in repeated five-fold CV~\cite{mostavi2020}, and GeneXNet applied deep learning to multi-tissue classification~\cite{khorshed2020}. A deep network interpreted with Shapley values identified 976 tissue-of-origin signature genes and exceeded 97\% accuracy on a random test split and 91\% on an independent ICGC cohort~\cite{divate2022}. Others were trained on TCGA and related compendia and tested on independent clinical cohorts, mainly to identify the origin of cancers of unknown primary~\cite{grewal2019,zhao2020,vibert2021}. Ensembles have been used as well: Mohammed et al.\ stacked one-dimensional convolutional networks on LASSO-selected genes for five cancer types and reported about 99\% accuracy on a held-out 30\% split~\cite{mohammed2021}. Third, expression signatures of specific cancers range from the 50-gene PAM50 predictor of breast-cancer subtypes~\cite{parker2009} to tumour versus normal markers derived from TCGA. The latter are complicated by the fact that tissue adjacent to a tumour differs from healthy tissue~\cite{aran2017}.

Most such classifiers are developed and evaluated within a single cohort, TCGA. In this setting, four questions are easy to overlook. First, once a strong gene panel has been selected, does a more complex model still help, or does it only add cost, and does the answer depend on how many genes the panel keeps? Ensembles such as stacking~\cite{wolpert1992} and hierarchical schemes that first assign a broad tumour group~\cite{silla2011} are attractive, but their benefit over a well-regularised linear model is rarely tested with rules fixed in advance. Evidence from general tabular data also shows that more complex models do not always win: in independent benchmarks, recently proposed deep networks did not outperform tuned tree ensembles~\cite{shwartzziv2022,grinsztajn2022}. Second, TCGA samples were contributed by hundreds of tissue source sites, and site is confounded with cancer type. Batch effects are a known source of optimism in high-throughput data~\cite{leek2010}; in TCGA RNA-seq, library size, tumour purity and sequencing plate are major sources of unwanted variation~\cite{molania2023}. Site signatures are also detectable in TCGA histology slides and inflate the accuracy of deep-learning models trained on them~\cite{howard2021,dehkharghanian2023}. Yet random splits leave samples from the same site on both sides of the split. The TCGA cancer-type classifiers cited above mostly report accuracy from random splits of samples~\cite{li2017,mostavi2020,mohammed2021,divate2022}. To our knowledge, none has been evaluated with folds grouped by tissue source site while repeating the gene selection inside each fold, so how much of their accuracy is tied to the contributing site is unknown. Third, does a panel transfer to an independent cohort processed with a different pipeline, where some of its genes may not even be measured? Fourth, small tumour versus normal marker panels are often ranked by differential expression on the whole cohort and scored on the same patients. Such panels look better than they are~\cite{ambroise2002}, and whether they are specific to early-stage disease is rarely checked.

Table~\ref{tab:related} summarises how representative classifiers of cancer type were validated. Most estimated accuracy from random splits of TCGA, and several added external cohorts, in some of which accuracy fell. Evaluation with sites kept apart has so far been reported for TCGA digital histology~\cite{howard2021}, not for these transcriptomic classifiers.

\begin{table}[!htbp]
\caption{Representative transcriptomic classifiers of cancer type and how they were validated. Validation and site columns describe what the studies report; ``not reported'' means that no site-grouped evaluation was described. TOO, tissue of origin; CUP, cancer of unknown primary; CV, cross-validation; ICGC, International Cancer Genome Consortium.\label{tab:related}}
\scriptsize
\renewcommand{\arraystretch}{1.25}\begin{tabularx}{\linewidth}{>{\raggedright\arraybackslash}p{1.6cm}LLLL>{\centering\arraybackslash}p{1.1cm}>{\centering\arraybackslash}p{1.3cm}L}
\toprule
\textbf{Study} & \textbf{Data and classes} & \textbf{Gene selection} & \textbf{Classifier} & \textbf{Validation} & \textbf{Site-grouped} & \textbf{External cohort} & \textbf{Open issue}\\
\midrule
Li et al.~\cite{li2017} & TCGA, 9096 tumours, 31 types & GA/kNN gene subsets & $k$-nearest neighbours & random 75/25 split & not reported & no & random split; READ largely indistinguishable from COAD\\
Grewal et al.~\cite{grewal2019} & TCGA and other public data; 40 cancer types and 26 normal tissues & whole transcriptome & ensemble of neural networks & independent clinical cohort (primary, metastatic, CUP) & not reported & yes & site dependence within training data not assessed\\
Mostavi et al.~\cite{mostavi2020} & TCGA, 10,340 tumours and 713 normals, 33 types + normal & mean/SD filter (7091 genes) & convolutional networks & repeated 5-fold CV & not reported & no & random splits only\\
Zhao et al.~\cite{zhao2020} & TCGA and ICGC, 18,217 primary tumours, 32 types & genes ordered by chromosome position & 1D Inception network (tuned) & CV; TCGA metastases; two clinical RNA-seq cohorts & not reported & yes & accuracy fell in clinical cohorts (87\%, 72\%)\\
Mohammed et al.~\cite{mohammed2021} & TCGA, 5 types & LASSO & stacked 1D convolutional networks & held-out 30\% split & not reported & no & five organ-distinct types; random split\\
Vibert et al.~\cite{vibert2021} & reference compendium, 20,918 samples; 39 cancer types, 55 normal tissues & --- & variational autoencoder & reference data; 48 CUP patients & not reported & yes & TOO identified in 79\% of CUP patients, whose origin is unverifiable\\
Divate et al.~\cite{divate2022} & TCGA, 14,237 tumours, 37 classes & expression filter; 976 SHAP-ranked genes & deep neural network & random 70/10/20 splits; 5-fold CV & not reported & yes (ICGC, 2085) & accuracy 97\% (TCGA) vs.\ 91\% (ICGC)\\
Howard et al.~\cite{howard2021} & TCGA digital histology & --- & deep learning on slides & site-preserved CV & yes & --- & histology, not expression; site signatures bias accuracy\\
This study & TCGA, 10,469 tumours, 33 types; SCAN-B & quality-aware hybrid Boruta, nested in folds & LR, 4 learners, 3 ensembles & random-split and site-grouped CV (5 assignments); locked test set & yes & yes (subtype panel) & site signal not attributable to technical vs.\ biological sources\\
\bottomrule
\end{tabularx}
\end{table}

In a previous study~\cite{paper1}, we examined the Boruta all-relevant feature selector~\cite{kursa2010} on TCGA RNA-seq and proposed a quality-aware hybrid selector. It filters genes whose expression follows sample quality within cancer type, applies category-aware ANOVA pre-selection across RNA categories, and runs Boruta at unlimited tree depth. Here we take that selector's panels as given and use them for two purposes: an ensemble classifier of cancer type and cancer-specific gene signatures. We address these four questions with four aims. (i)~We compare five learners and three ensembles under pre-declared decision rules, with the 1000-gene panel and a minimal 100-gene panel. (ii)~We measure site dependence by repeating the whole selection in cross-validation grouped by tissue source site, under several assignments of sites to folds. (iii)~We test transfer of a breast-cancer subtype panel to the population-based SCAN-B cohort~\cite{saal2015,brueffer2018}. (iv)~We derive small (5--20-gene) stage~I tumour versus adjacent-normal panels with differential expression and pathway enrichment, evaluated in patients held out from all model building. Every decision rule, threshold and test-set use was fixed before the corresponding results were seen; analyses added afterwards are reported as exploratory.

Rather than introducing a new classification or gene selection algorithm, this study evaluates how panel size, dependence on the contributing site and cross-cohort compatibility affect the reliability of TCGA-derived classifiers and gene signatures. Our contributions are as follows. (1)~Under pre-declared rules and with gene selection nested in every training fold, no ensemble improved on a well-calibrated logistic regression by the pre-declared margin with a 1000-gene panel; stacking helped consistently only with the minimal 100-gene panel. (2)~We evaluated a TCGA pan-cancer classifier with folds grouped by tissue source site, rerunning the selector inside each fold and repeating the analysis over five assignments of sites to folds; the optimism of random splits was concentrated in colon and rectal cancer, the same pair that dominates the remaining errors. (3)~A TCGA subtype panel transferred to a population cohort with accuracy comparable to the baseline panel and close to the PAM50 genes; its non-coding genes were largely absent from the external annotation, so only its protein-coding core was validated. (4)~For nine cancer types, we derived panels of five genes that distinguish stage~I tumours from adjacent normal tissue; the genes were selected inside every training fold, the panels were tested in held-out patients and in later-stage tumours, and we report how stable the selected genes were across folds.

\section{Materials and Methods}

\begin{figure}[!htbp]
\includegraphics[width=\linewidth]{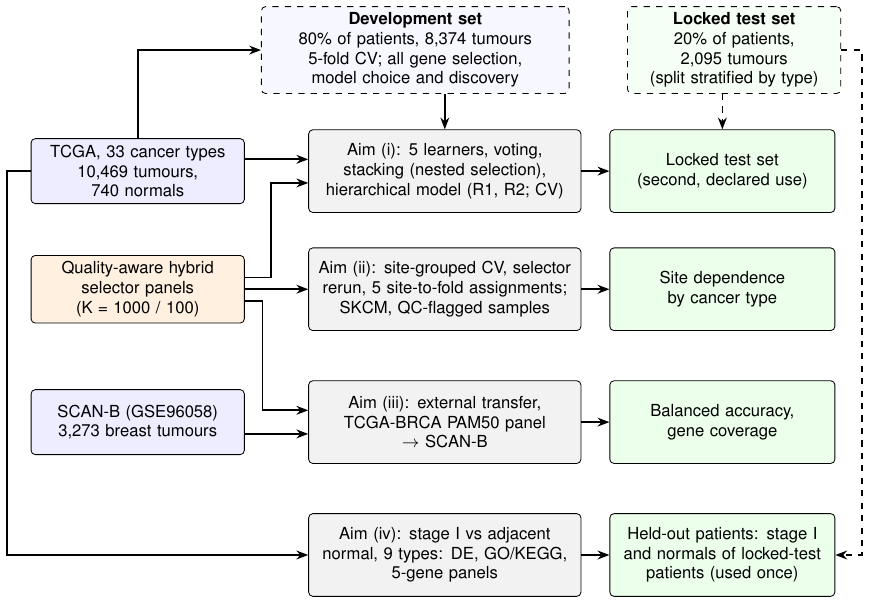}
\caption{Study design. Patients were split once, stratified by cancer type, into a development set (80\%), used for all gene selection, model choice and marker discovery, and a locked test set (20\%), whose tumours were used for the declared second evaluation of the cancer-type models and whose stage~I tumours and normals formed the held-out patients of aim (iv). The gene panels of the quality-aware hybrid selector feed four analyses, one per aim: (i)~model comparison under pre-declared rules; (ii)~robustness checks with the selector rerun inside site-grouped folds; (iii)~external transfer of a breast-cancer subtype panel to SCAN-B; (iv)~stage~I tumour vs.\ adjacent-normal signatures, with patients held out from all model building. QC, quality control; CV, cross-validation; DE, differential expression.\label{fig:design}}
\end{figure}

Figure~\ref{fig:design} summarises the workflow. The gene panels of the quality-aware hybrid selector feed one analysis per aim (i)--(iv), each with its own evaluation design.

\subsection{Data and Cohorts}

We used the same cohort as in our previous study~\cite{paper1}, which we describe briefly here. RNA-seq gene expression (STAR--Counts workflow, GENCODE v36 annotation~\cite{frankish2021}) for all 33 TCGA projects was downloaded from the NCI Genomic Data Commons~\cite{grossman2016}. The class of each tumour is its TCGA project (cancer type). One primary tumour per patient was kept (blood-derived for LAML; metastatic samples were kept for SKCM, as most SKCM tumours in TCGA are metastases). Expression was represented as log$_2$(TPM~+~1), and raw counts were kept for differential expression analysis. Tumours whose number of detected genes (TPM~>~1) was more than three median absolute deviations below their cancer-type median and below 70\% of that median were excluded (19 tumours). A further 253 tumours, flagged by a looser rule (library size or detected genes more than three median absolute deviations below the median), were kept. In total, 10,469 tumours (33~types) and 740 solid-tissue normal samples were kept. Twenty percent of patients, stratified by cancer type, formed a locked test set (2095 tumours). The remaining 8374 tumours were the development set, used with five-fold stratified cross-validation (CV). The cohorts used for each analysis are listed in Table~\ref{tab:cohorts}.

\begin{table}[!htbp]
\caption{Cohorts and tasks.\label{tab:cohorts}}
\begin{tabularx}{\linewidth}{lLLcc}
\toprule
\textbf{Task} & \textbf{Training data} & \textbf{Evaluation} & \textbf{Classes} & \textbf{Section}\\
\midrule
Cancer type & TCGA development, 8374 tumours & 5-fold CV; locked test set (2095) & 33 & \ref{sec:m-models}, \ref{sec:m-robust}\\
Breast cancer subtype & TCGA-BRCA development, 797 tumours & SCAN-B (GSE96058), 3273 tumours & 5 (PAM50) & \ref{sec:m-external}\\
Stage I vs.\ adjacent normal & 9 TCGA types, development patients & patient-grouped CV; held-out patients & 2 per type & \ref{sec:m-stage}\\
\bottomrule
\end{tabularx}
\end{table}

\subsection{Gene Panels from the Quality-Aware Hybrid Selector}

The quality-aware hybrid selector~\cite{paper1} has four steps. (1)~An expression filter keeps genes with TPM~>~1 in at least 20 training samples. (2)~A quality filter removes genes whose within-cancer-type Pearson correlation with log$_{10}$ detected genes exceeds 0.3 in absolute value. (3)~Category-aware ANOVA F pre-selection keeps 3000 candidates, with quotas for protein-coding, lncRNA, miRNA and other genes. (4)~Boruta~\cite{kursa2010} is run with 300 trees of unlimited depth, 50 iterations and $\alpha = 0.05$, with three seeds, and genes are ranked by seed-averaged hit rate. A panel of size $K$ consists of the $K$ top-ranked genes. All four steps run inside each training fold. Here we reused, without change, the fold-wise rankings of the CV folds and the ranking fitted on the whole development set, and reran the selector only where a design required new training sets (nested stacking and site-grouped CV). The main panel size was $K = 1000$, chosen for logistic regression by the one-standard-error rule~\cite{hastie2009}; a minimal panel of $K = 100$ genes was a secondary analysis, to test whether ensembles help when few genes are available. For the breast-cancer subtype task, we reused the PAM50 rankings of the same selector and their panel sizes, with labels from the TCGA PanCancer Atlas~\cite{hoadley2018}, obtained through cBioPortal~\cite{cerami2012,gao2013} and defined by the PAM50 genes~\cite{parker2009,tcga2012}. Supplementary Methods~S1 describes the selector in full, including the choice of panel size, and lists the cancer types in each hard group of the hierarchical model (Section~\ref{sec:m-models}).

\emph{Selection stability.} We measured how reproducible the selected genes were from the saved selector runs, without refitting. Stability across Boruta seeds was the Jaccard index of the top-$K$ genes of the single-seed rankings on the same training data. Stability under resampling was the pairwise Jaccard index and Nogueira's stability measure~\cite{nogueira2018} of the top-$K$ genes across training sets: the five random-split training folds, the 25 inner training sets of the nested stacking design (Section~\ref{sec:m-models}) and the 25 site-grouped training folds (Section~\ref{sec:m-robust}). For each gene of the final panel, we counted how often it was in the top $K$ of these 55 resampled rankings. The chance level is the expected Jaccard index of two random sets of $K$ out of the 3000 candidates, $K/(6000 - K)$.

\subsection{Learners and Ensembles}\label{sec:m-models}

Five base learners were fitted with fixed settings and no tuning:
\begin{itemize}
\item L2 logistic regression (LR; $C = 0.1$, balanced class weights, standardised inputs);
\item LightGBM~\cite{ke2017} (300 trees, learning rate 0.05, 31 leaves, balanced class weights, deterministic mode);
\item a random forest~\cite{breiman2001} (500 trees, $\sqrt{p}$ features per split, balanced class weights per bootstrap sample);
\item a linear support vector machine ($C = 0.01$, balanced class weights, standardised inputs) with sigmoid calibration in three internal folds;
\item a multilayer perceptron (one hidden layer of 256 units, L2 penalty $10^{-3}$, standardised inputs, early stopping, at most 300 epochs).
\end{itemize}
These settings were fixed in advance and not tuned, so the comparisons concern these configurations rather than the learning algorithms in general.
The learners were implemented in scikit-learn~\cite{pedregosa2011} and LightGBM. Soft voting averaged the five learners' class probabilities. Stacking~\cite{wolpert1992} trained a multinomial LR meta-learner ($C = 1$, balanced class weights) on out-of-fold probabilities from five stratified inner folds within every outer training fold. Gene selection was nested: the complete selector was rerun inside each of the 25 inner training sets (75 Boruta runs), so the inner validation samples never took part in selecting the genes behind their out-of-fold probabilities. The base learners for the outer validation fold were fitted on the whole outer training fold with that fold's panel. A non-nested variant, in which the inner folds reused the outer fold's panel, is reported for comparison (Table~S2). The hierarchical model used seven pre-defined groups of similar cancer types that are hard to separate (hard groups) (colorectal; upper gastrointestinal adenocarcinoma; squamous; lung adenocarcinoma; kidney; glioma; uterine), defined following the pan-cancer analysis~\cite{hoadley2018}. It combined the flat model's group probability with a specialist model trained within the group: $P(c) = \sum_g P(g)\,P(c \mid g)$. The specialist was soft voting of the five learners refitted on the group's training samples. Oesophageal carcinoma (ESCA) tumours were assigned to a group by histology; because ESCA adenocarcinomas and squamous tumours belong to different groups, the flat model's ESCA probability was split between the two groups in proportion to their ESCA training samples.

Models were compared by balanced accuracy (primary measure~\cite{brodersen2010}), macro-F1, log loss, expected calibration error (ECE; 15 bins~\cite{naeini2015}) and the within-group confusion rate. The within-group confusion rate is the share of hard-group samples predicted as another member of their own group. Two decision rules were fixed before any model was evaluated. Rule R1: stacking replaces the best single learner only if its mean CV gain is at least 0.005, the gain is positive in at least four of five folds, and an exact McNemar test~\cite{mcnemar1947,dietterich1998} on the pooled out-of-fold predictions gives $p < 0.05$. Rule R2: the hierarchical model replaces the model chosen by R1 only if within-group confusion falls by at least 10\% (relative) and balanced accuracy drops by no more than 0.002. The locked test set had been opened once before, in our previous study~\cite{paper1}, to evaluate the same LR model and panel. Its second use here was declared in advance and made only after both rules had been recorded; no design decision of the present study depended on test results. Because the test set is no longer untouched, it confirms the reference model rather than providing an independent estimate for model selection: on the test set, the LR model of that study is the reference, and all other models are secondary results. On the test set, stacking used the non-nested meta-learner fitted on the whole development set.

\subsection{Robustness Checks}\label{sec:m-robust}

\emph{Site effects.} TCGA issues tissue source site (TSS) codes per contributing institution and study. In the development set, every one of the 672 codes contributed tumours of a single cancer type, and no code contributed both colon and rectal tumours (24 COAD-only and 13 READ-only codes). Under random splits, the contributing collection is therefore a perfect proxy for the label, and any signal associated with it can raise accuracy, whether technical (sample handling, processing, batch) or biological (patient population, sampling practice). Site grouping cannot tell these sources apart; it measures how much accuracy depends on seeing the same collections in training and validation. To remove this shortcut, development samples were split into five folds grouped by TSS code and stratified by cancer type, so that no code appears in both the training and the validation fold (StratifiedGroupKFold in scikit-learn; each sample without a recorded code formed its own group). The complete hybrid selector was rerun inside every site-grouped training fold (15 Boruta runs), so no gene selection used validation sites. Balanced accuracy was averaged over the five site-grouped folds. Interpretation bands were fixed in advance: a drop in balanced accuracy relative to random-split CV of at most 0.01 is small, 0.01--0.03 moderate and larger than 0.03 substantial. Cancer types from a single site cannot be evaluated under site grouping and were listed separately. We also pooled the out-of-fold predictions of both CV schemes, which predict every development sample once, and compared them sample by sample: the drop in balanced accuracy with a paired stratified bootstrap 95\% CI~\cite{efron1993}, and per cancer type the change in recall with an exact McNemar test and Benjamini--Hochberg adjustment over types~\cite{benjamini1995}. Because StratifiedGroupKFold assigns sites to folds greedily and the assignment depends on its random seed, we repeated the site-grouped CV, including gene selection, with four further assignments of sites to folds (seeds 1--4; 60 additional Boruta runs).

\emph{Sample type and quality.} From the out-of-fold LR predictions, we compared recall for primary and metastatic SKCM tumours, with exact binomial confidence intervals. We also compared accuracy for development samples flagged by quality control but kept with that for unflagged development samples. Finally, we classified the 19 QC-excluded tumours, which were never used for training.

\subsection{External Transfer to SCAN-B}\label{sec:m-external}

GSE96058 contains RNA-seq profiles of 3273 primary breast tumours from the population-based SCAN-B initiative, on HiSeq~2000 and NextSeq~500 instruments, with PAM50 subtype calls~\cite{saal2015,brueffer2018}. Technical replicates were removed. The 797 TCGA-BRCA development tumours were the training set. Genes were matched by symbol, and symbols that were ambiguous in either cohort were not used. Each panel was filled by walking down its ranking until $K$ genes present in SCAN-B were found. Each cohort was $z$-scored per gene within itself: TCGA on log$_2$(TPM~+~1) and SCAN-B on its log$_2$ FPKM scale. SCAN-B labels were used only for evaluation. The primary comparison, fixed in advance, was the quality-aware panel vs.\ the baseline panel of our previous study, which was selected by the same pipeline without the quality filter, with LR at $K = 50$, the size chosen by the one-standard-error rule. It used a paired stratified bootstrap 95\% CI of the difference in balanced accuracy (2000 resamples~\cite{efron1993}) and an exact McNemar test; a difference would be claimed only if the CI excluded zero. The depth-5 panel, the 50 PAM50 genes and LightGBM were secondary analyses. Because SCAN-B subtypes are themselves PAM50 calls, the PAM50 genes are an upper reference rather than a fair competitor. Finally, we compared how much each panel depends on sample quality, as the difference in balanced accuracy between SCAN-B samples above and below the subtype median of detected genes. In post hoc analyses, we compared the HiSeq~2000 and NextSeq~500 tumours (balanced accuracy with stratified bootstrap CIs, subtype mix, detected genes and clinical characteristics), tabulated confusion matrices and classified the genes that could not be matched by RNA category.

\subsection{Stage I Tumour vs.\ Adjacent-Normal Markers}\label{sec:m-stage}

\emph{Eligibility and splits.} Cancer types with at least 30 normal samples and at least 30 stage~I tumours (AJCC pathological stage) in total were eligible. Patients in the locked test set of the cancer-type task were kept out of all marker discovery. Their stage~I tumours and normals formed a held-out set, used once and only where it contained at least five samples of each kind.

\emph{Differential expression and enrichment.} Differential expression (DE) used PyDESeq2~\cite{muzellec2023,love2014} on raw counts with design $\sim$condition. Genes needed TPM~>~1 in at least 10 of the samples being compared. DE genes had Benjamini--Hochberg~\cite{benjamini1995} adjusted $p < 0.05$ and an apeGLM-shrunken~\cite{zhu2019} $|\log_2 \mathrm{FC}| \geq 1$. A sensitivity analysis with design $\sim$patient~+~condition used paired patients where at least 10 pairs existed. Enrichment of up- and down-regulated genes in GO Biological Process 2023~\cite{go2023} and KEGG 2021~\cite{kanehisa2000} (Enrichr libraries~\cite{kuleshov2016}, via GSEApy~\cite{fang2023}) used a hypergeometric test. The background was the genes expressed in that cancer type, and only terms with 5--500 genes and adjusted $p < 0.05$ were counted.

\emph{Classifiers.} Stage~I vs.\ normal classifiers were evaluated in five-fold CV grouped by patient, so that a tumour and its paired normal stayed together. Inside every training fold, genes were ranked either by DE $p$-value (among genes with $|\log_2\mathrm{FC}| \geq 1$) or by the quality-aware hybrid selector adapted to the binary task: the quality filter used correlations within tumour and within normal samples, the 3000 candidates were the genes with the highest ANOVA F without category quotas, and Boruta was run with a single seed. LR was fitted on the top $K = 5$, 10 or 20 genes. The pre-declared rule chose per cancer type the selector with the higher mean CV ROC AUC at $K = 10$; if the two were within 0.005, it chose DE, the simpler selector. It then took the smallest $K$ within 0.01 of that selector's best AUC. The chosen panel was refitted on all development patients and applied once to the held-out patients. To test stage specificity, it was also applied to development stage~II and stage~III--IV tumours, which were never used for fitting. Panel stability was the Jaccard index between the top five genes of each CV training fold and the final panel.

\subsection{Statistics and Software}

Confidence intervals for balanced accuracy were stratified bootstrap percentile intervals (2000 resamples). Paired model comparisons used the exact McNemar test on per-sample correctness. Proportions used exact binomial (Clopper--Pearson) intervals.

\emph{Status of the analyses.} Primary analyses, with designs and decision rules fixed in advance, were: rule R1 at $K = 1000$ and rule R2; the site-grouped drop with its interpretation bands; the quality-aware vs.\ baseline comparison on SCAN-B; and the stage~I panel rule with its single held-out evaluation. Secondary analyses, also declared in advance, were the 100-gene panel, LightGBM, the depth-5 and PAM50 panels on SCAN-B, and the later-stage tumours. The following were post hoc and are exploratory: the site analysis without COAD, READ and LAML, the further site-to-fold assignments, the per-type recall tests, selection stability and the SCAN-B platform comparison.

Analyses used Python 3.10.12 with NumPy 2.2.6~\cite{harris2020}, pandas 2.3.3, SciPy 1.15.3~\cite{virtanen2020}, scikit-learn 1.7.2~\cite{pedregosa2011}, LightGBM 4.7.0~\cite{ke2017}, PyDESeq2 0.5.2~\cite{muzellec2023}, GSEApy 1.3.1~\cite{fang2023} and Matplotlib 3.10.8~\cite{hunter2007}, on a 47-core server without a GPU. Exact versions of all packages are listed in the code repository.

\section{Results}

\subsection{A Well-Regularised Linear Model Is Hard to Beat}

\begin{figure}[!htbp]
\centering
\includegraphics[width=\linewidth]{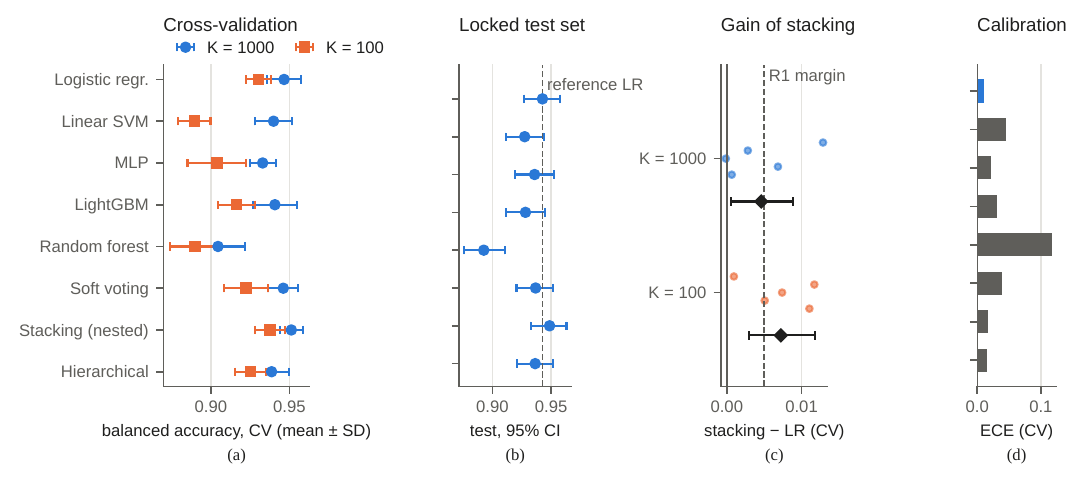}
\caption{Model comparison. (\textbf{a}) CV balanced accuracy (mean $\pm$ SD) with the 1000- and 100-gene panels; stacking with gene selection nested in the inner folds. (\textbf{b}) Locked test set, 1000 genes, 95\% bootstrap CIs; dashed line, reference LR; stacking uses the non-nested meta-learner. (\textbf{c}) Gain of nested stacking over LR in each CV fold (dots), with the mean and paired bootstrap 95\% CI (diamond); dashed line, pre-declared margin of rule R1. (\textbf{d}) Expected calibration error in CV, 1000 genes (stacking, non-nested variant); LR highlighted.\label{fig:models}}
\end{figure}

In CV with the 1000-gene panel, LR was the best single learner (balanced accuracy 0.947~$\pm$~0.011; Table~\ref{tab:models}, Figure~\ref{fig:models}a). It also had the lowest calibration error (ECE 0.010) and log loss (0.137). With nested gene selection, stacking reached 0.951, a mean gain of 0.0046 (paired bootstrap 95\% CI 0.0005--0.0089) that was positive in four of five folds (McNemar $p = 0.037$). The gain was below the pre-declared margin of 0.005, so rule R1 kept LR (Figure~\ref{fig:models}c). With $K = 100$, stacking improved on LR in all five folds by 0.0072 on average (0.938 vs.\ 0.930; 95\% CI 0.0030--0.0118; $p < 0.001$), a secondary result. Nesting the selection changed little: the non-nested variant gave 0.950 and 0.936, within 0.002 of the nested values (McNemar $p = 0.70$ and 0.18; Table~S2), and the inner-fold panels shared most of their genes with the outer-fold panels (mean Jaccard index 0.82). LightGBM was the most costly learner, taking on average 17.6 min per outer-fold fit at $K = 1000$, compared with 0.15~min for LR.

On the locked test set, LR at $K = 1000$ reached a balanced accuracy of 0.943 (95\% CI 0.927--0.958), identical to the value reported in our previous study~\cite{paper1}, and its prediction for every test tumour was the same as before. This confirms that the models were reproduced exactly. Stacking, with the non-nested meta-learner, was again highest (0.949), but its advantage over LR was not significant: +0.006, 95\% CI $-0.002$ to $+0.015$, McNemar $p = 0.15$. LR remained the best-calibrated model on the test set (ECE 0.007; Table~\ref{tab:models}), as in CV (Figure~\ref{fig:models}d).

\begin{table}[!htbp]
\caption{Model comparison, cancer type (33 classes). CV, mean $\pm$ SD over five folds; test, locked test set (2095 tumours) with 95\% bootstrap CI. ECE, expected calibration error.\label{tab:models}}
\begin{tabularx}{\linewidth}{lCCCCC}
\toprule
\textbf{Model} & \textbf{CV, $K = 1000$} & \textbf{CV, $K = 100$} & \textbf{Test, $K = 1000$} & \textbf{Test log loss} & \textbf{Test ECE}\\
\midrule
Logistic regression (reference) & 0.947 $\pm$ 0.011 & 0.930 & 0.943 (0.927--0.958) & 0.126 & 0.007\\
Linear SVM & 0.940 $\pm$ 0.012 & 0.890 & 0.928 (0.911--0.944) & 0.180 & 0.046\\
MLP & 0.933 $\pm$ 0.008 & 0.904 & 0.936 (0.919--0.952) & 0.149 & 0.018\\
LightGBM & 0.941 $\pm$ 0.014 & 0.916 & 0.928 (0.912--0.945) & 0.197 & 0.027\\
Random forest & 0.905 $\pm$ 0.018 & 0.890 & 0.893 (0.876--0.911) & 0.287 & 0.109\\
Soft voting & 0.946 $\pm$ 0.010 & 0.922 & 0.937 (0.921--0.952) & 0.145 & 0.039\\
Stacking (nested)$^{1}$ & 0.951 $\pm$ 0.008 & 0.938 & 0.949 (0.933--0.963) & 0.140 & 0.018\\
Hierarchical & 0.939 $\pm$ 0.011 & 0.925 & 0.936 (0.921--0.952) & 0.138 & 0.021\\
\bottomrule
\end{tabularx}
\noindent{\footnotesize $^{1}$ CV values with gene selection nested in the inner folds; the test-set values use the non-nested meta-learner fitted on the whole development set. Non-nested CV values: 0.950 $\pm$ 0.009 ($K = 1000$) and 0.936 ($K = 100$).}
\end{table}

\subsection{Errors Concentrate in Two Biologically Close Pairs}

Many of the remaining errors were confusions within hard groups (Figure~\ref{fig:sites}a). The within-group confusion rate was 0.19 for colon vs.\ rectal adenocarcinoma and 0.10 for stomach vs.\ oesophageal adenocarcinoma. It was 0.034 for kidney tumours, 0.039 for LUAD vs.\ LUSC and below 0.03 for glioma, squamous and uterine tumours. The hierarchical model did not reduce within-group confusion (0.047 vs.\ 0.045 for LR) and lowered balanced accuracy by 0.008, so rule R2 kept the flat LR model. On the test set, its difference from LR was $-0.006$ (95\% CI $-0.015$ to $+0.002$).

\subsection{Site-Grouped Validation Reveals Site-Associated Optimism in Colorectal Cancer}

\begin{figure}[!htbp]
\centering
\includegraphics[width=\linewidth]{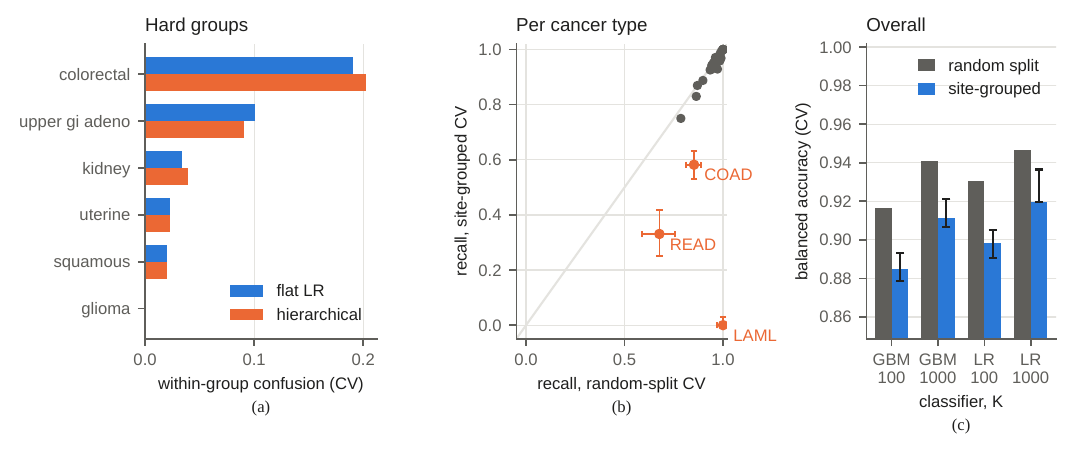}
\caption{Hard groups and site effects. (\textbf{a}) Within-group confusion of the flat LR and the hierarchical model (CV; the single-type lung group is omitted). (\textbf{b}) Recall per cancer type under random-split vs.\ site-grouped CV (LR, 1000 genes; pooled out-of-fold predictions); types losing more than 0.1 are labelled, with exact 95\% CIs. LAML came from a single site and cannot be tested under site grouping. (\textbf{c}) Balanced accuracy under both CV schemes (fold means); error bars, range of the site-grouped value over five site-to-fold assignments; GBM, LightGBM.\label{fig:sites}}
\end{figure}

With sites kept apart and the selector rerun inside each fold, balanced accuracy of LR at $K = 1000$ fell from 0.947 to 0.920, a moderate drop of 0.027 under our pre-declared bands (Figure~\ref{fig:sites}c). LightGBM fell by 0.029, and both learners fell by about 0.032 with 100 genes. LAML came from a single site and could not be tested; pooling the out-of-fold predictions without it, the drop was 0.026 (95\% CI 0.021--0.033). Only three types changed significantly (adjusted McNemar $p < 0.05$; Table~S3, Figure~S2): LAML, whose recall fell from 1.00 to 0 because its single site left no LAML tumours for training in its fold; rectal adenocarcinoma, whose recall fell from 0.68 to 0.33; and colon adenocarcinoma, from 0.85 to 0.58 (Figure~\ref{fig:sites}b). The colon and rectal errors were almost all confusions with each other: 151 colon tumours were called rectal and 89 rectal tumours colon. Every other testable cancer type changed by less than 0.05. READ was also concentrated in few collections: one of its 13 codes contributed 44\% of its development tumours, so a site-grouped fold could remove a large share of the rectal training data at once, and part of the drop may reflect this. Concentration alone did not cause large drops, however: one code contributed 90\% of ACC tumours, yet ACC recall did not change (0.951 under both schemes). In an exploratory analysis that excluded COAD, READ and LAML, the drop was 0.008 (95\% CI 0.003--0.013), within the small band. LightGBM was more site-dependent: without these three types it lost 0.030 (0.022--0.039).

The result did not depend on how sites were assigned to folds. Under four further assignments, generated with different random seeds, the fold-mean drop for LR at $K = 1000$ ranged from 0.010 to 0.027 (mean 0.021); the original assignment gave the largest value (Figure~\ref{fig:sites}c, Table~S4). Without LAML, COAD and READ, the pooled drop was 0.003--0.008 under every assignment, and LightGBM was more site-dependent than LR under all five. Rectal tumours lost recall under every assignment (0.25--0.39), whereas the loss for colon tumours ranged from 0 to 0.27, depending on which collections were held out. Because every TSS code carries a single cancer type, these drops measure the optimism of sample-level random splitting that is associated with the contributing site. They do not identify its cause, which may be technical, such as processing and batch, or biological, such as differences in the patients and samples each collection contributed.

The other two robustness checks showed only small effects. Metastatic SKCM tumours were classified as well as primary ones (recall 0.969, $n = 286$, vs.\ 0.965, $n = 86$). Development samples flagged by QC but kept were slightly less accurate than unflagged ones (accuracy 0.947 vs.\ 0.958, $n = 207$ flagged). Of the 19 QC-excluded tumours, 18 were classified correctly; the one error was a KIRC tumour called KICH.

Gene selection itself was stable (Table~S5, Figure~S1). On the same training data, the three Boruta seeds gave top-$K$ panels with mean Jaccard indices of 0.88--0.93 ($K = 50$--1000), and their sets of confirmed genes 0.94. Across the five random-split training folds and the 25 inner training sets, the overlap was 0.74--0.83 (Nogueira's measure 0.85--0.91), far above the chance level of 0.008--0.20. Site-grouped training folds agreed less (0.64--0.72), consistent with the site dependence above. Of the final top-100 genes, 62 were in the top 100 of at least 90\% of the 55 resampled rankings, and 49 in all of them. Protein-coding genes were selected more consistently than lncRNAs and pseudogenes: of the final top-1000 genes, 75\% of the protein-coding genes but 63\% of the lncRNAs and 52\% of the other genes reached this 90\% level.

\subsection{The Subtype Panel Transfers to an Independent Cohort}

\begin{figure}[!htbp]
\centering
\includegraphics[width=\linewidth]{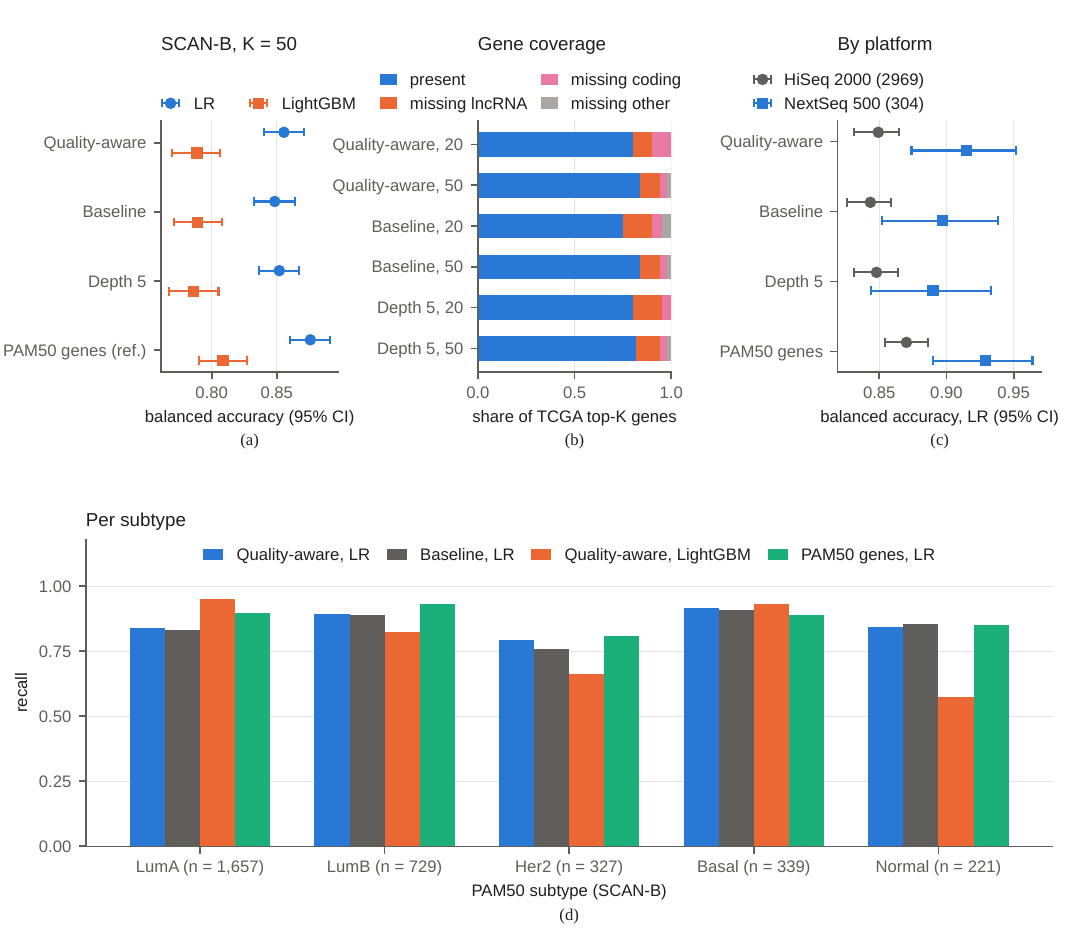}
\caption{External transfer, TCGA-BRCA to SCAN-B. (\textbf{a}) Balanced accuracy on SCAN-B with 50 genes (95\% bootstrap CI); the PAM50 genes are an upper reference because SCAN-B subtypes are PAM50 calls. (\textbf{b}) Share of each panel's TCGA top-$K$ genes measured in SCAN-B, with the missing genes by RNA category. (\textbf{c}) Balanced accuracy by sequencing platform, LR with 50 genes (95\% bootstrap CIs; number of tumours in brackets). (\textbf{d}) Recall per subtype.\label{fig:external}}
\end{figure}

On 3273 SCAN-B tumours, the quality-aware panel with LR reached a balanced accuracy of 0.855 (95\% CI 0.840--0.871), compared with 0.848 for the baseline panel (Figure~\ref{fig:external}a). The pre-declared comparison did not show a difference (+0.007, 95\% CI $-0.003$ to $+0.017$, McNemar $p = 0.077$), so the two panels performed comparably. Both came within 0.03 of the 50 PAM50 genes (0.876), the upper reference. LR transferred much better than LightGBM, both for the Boruta panels (0.855 vs.\ 0.789) and for the PAM50 genes (0.876 vs.\ 0.809). Recall was highest for basal-like and luminal~B tumours and lowest for HER2-enriched tumours (0.79 with the quality-aware panel, 0.76 with the baseline panel; Figure~\ref{fig:external}d). With LightGBM, recall for normal-like tumours dropped to 0.57. The errors of the quality-aware panel resembled those of the PAM50 genes: luminal~A tumours called normal-like (169 tumours, vs.\ 95 with the PAM50 genes) or luminal~B, and HER2-enriched tumours called luminal~B (Table~S6, Figure~S3).

Of the top-20 and top-50 TCGA genes, 15--16 and 41--42 were present in SCAN-B. Every missing gene was absent from SCAN-B's gene list; none was excluded for an ambiguous symbol. The missing genes were mostly lncRNAs and other non-coding genes: of the six lncRNAs among the top 60 genes of the quality-aware ranking, five were missing, against three of 51 protein-coding genes (SGO1, NCBP2L and TESMIN, whose symbols may differ between annotation versions). The panels used on SCAN-B were therefore 96--100\% protein-coding (Figure~\ref{fig:external}b; Table~S6). Both panels performed slightly better on lower-quality samples (balanced accuracy above minus below the median of detected genes, $-0.021$ and $-0.030$), and the difference between their gaps had a CI that included zero (0.009, $-0.010$ to $0.028$).

Accuracy also differed between the two sequencing platforms. It was higher on the 304 NextSeq than on the 2969 HiSeq tumours for all 11 models (LR with the quality-aware panel: 0.915 vs.\ 0.849; difference 0.065, 95\% CI 0.022--0.106; Figure~\ref{fig:external}c). The gap was similar for the quality-aware and the baseline panel (difference between their gaps 0.012, $-0.014$ to 0.043) and appeared with the PAM50 genes as well (0.059). The two platforms had nearly identical subtype mix ($p = 0.96$), detected-gene counts and ER, HER2, grade, node and age distributions (Table~S6). The gap is therefore most likely associated with the sequencing platform, or with processing differences that accompanied it, rather than with the selected panels. Because only 304 tumours were sequenced on NextSeq, the size of the gap is uncertain, as its wide CI shows.

\subsection{Stage I Tumours Differ from Adjacent Normal Tissue Through a Shared Programme}

\begin{figure}[!htbp]
\centering
\includegraphics[width=\linewidth]{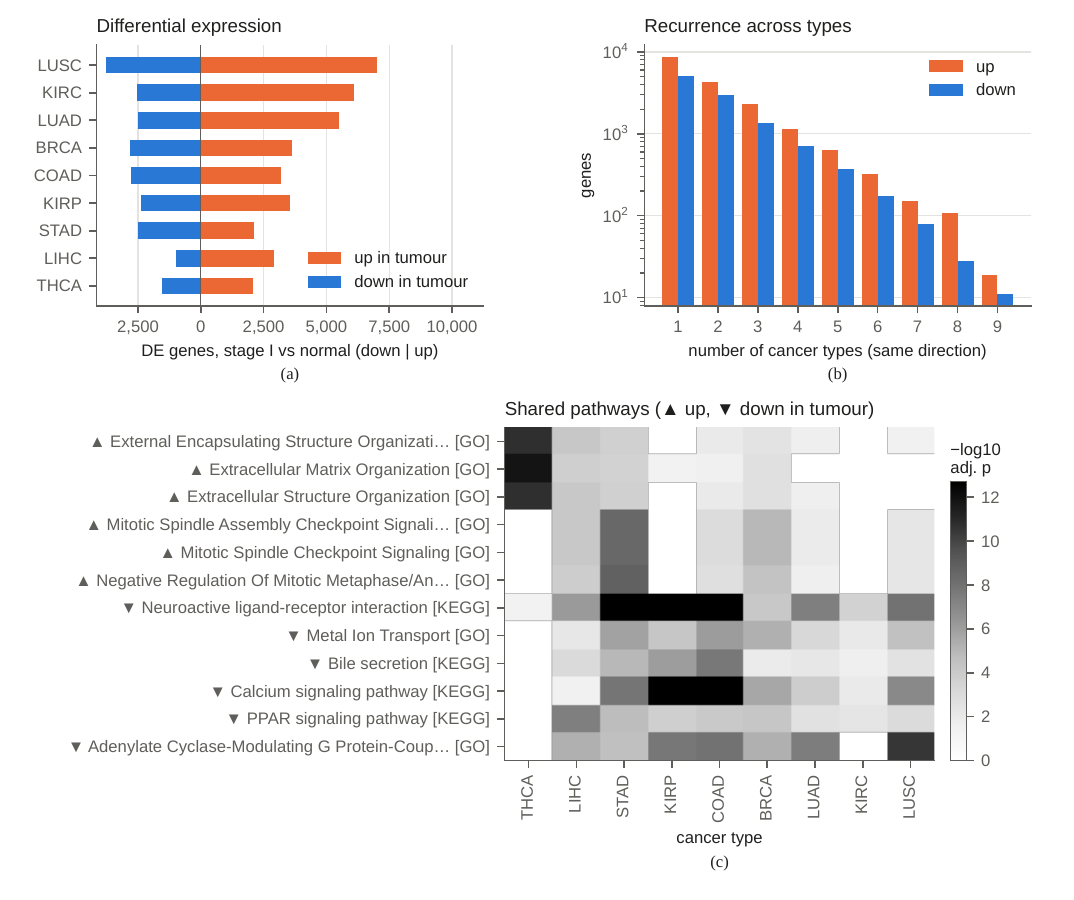}
\caption{Stage~I tumours vs.\ adjacent normal tissue (development patients). (\textbf{a}) DE genes per cancer type (adjusted $p < 0.05$, $|\log_2\mathrm{FC}| \geq 1$). (\textbf{b}) Number of cancer types in which a gene is DE in the same direction (log scale). (\textbf{c}) The GO Biological Process and KEGG terms significant in the most cancer types; shading, $-\log_{10}$ adjusted $p$.\label{fig:de}}
\end{figure}

Nine cancer types had at least 30 stage~I tumours and 30 adjacent normal samples in total and were therefore eligible for this analysis (BRCA, COAD, KIRC, KIRP, LIHC, LUAD, LUSC, STAD, THCA; Section~\ref{sec:m-stage}). Between 3615 (THCA) and 10,796 (LUSC) genes were differentially expressed between stage~I tumours and adjacent normal tissue (Figure~\ref{fig:de}a). Of these, 47--65\% were protein-coding and 16--29\% lncRNAs. In the seven types with at least 10 tumour--normal pairs, the paired-patient analysis overlapped the main results with Jaccard indices of 0.36--0.65, although the paired subsets had only 10--32 pairs. In total, 1911 genes changed in the same direction in at least five of the nine types (Figure~\ref{fig:de}b).

The shared pathways were consistent with known tumour biology (Figure~\ref{fig:de}c). Genes up-regulated in tumours were enriched for extracellular-matrix organisation (seven types), the mitotic spindle checkpoint and sister chromatid segregation (six types each). Down-regulated genes were enriched for signalling and secretory functions of the normal tissues: neuroactive ligand--receptor interaction in all nine types, calcium signalling, PPAR signalling and bile secretion in eight, and gastric acid secretion in seven.

\subsection{Five-Gene Panels Separate Tumours from Adjacent Normal Tissue from Stage I Onward}

\begin{figure}[!htbp]
\centering
\includegraphics[width=\linewidth]{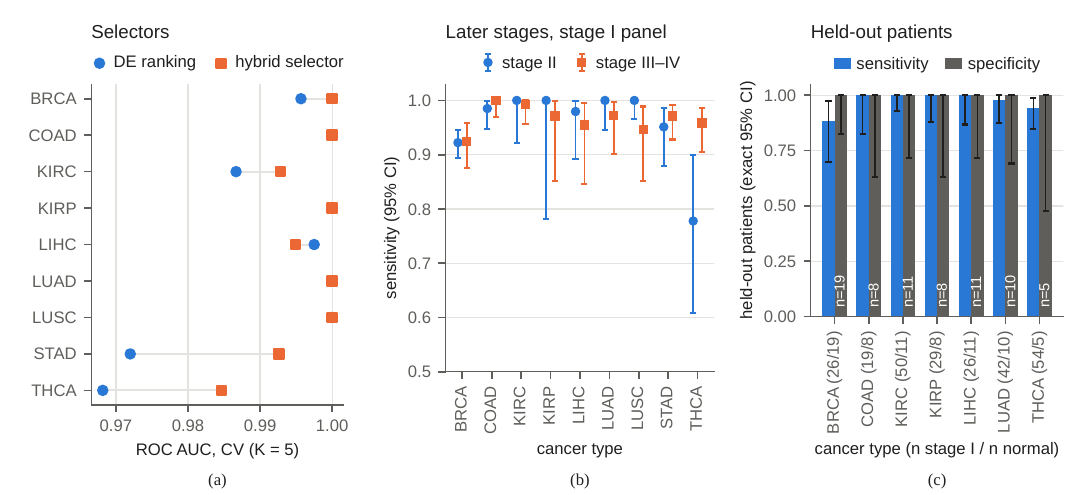}
\caption{Stage~I classifiers. (\textbf{a}) CV ROC AUC at $K = 5$ with genes ranked by DE or by the quality-aware hybrid selector. (\textbf{b}) Sensitivity of each stage~I panel for development stage~II and stage~III--IV tumours, never used for fitting (95\% exact CI). (\textbf{c}) Sensitivity and specificity in held-out patients with exact 95\% CIs (n stage~I / n normal; the number of normals is printed in each specificity bar).\label{fig:stage}}
\end{figure}

Five genes sufficed in every eligible type. ROC AUC in patient-grouped CV was 0.97--1.00 in every type (Figure~\ref{fig:stage}a), and the pre-declared rule chose $K = 5$ everywhere (Table~\ref{tab:stage}). These estimates are free of selection leakage: gene ranking was repeated inside every training fold, and the held-out patients were never seen by DE, enrichment, ranking or the panel-size rule. In the three types where the task was hardest, the hybrid selector gave a higher CV AUC than DE ranking at $K = 5$: STAD (0.993 vs.\ 0.972), THCA (0.985 vs.\ 0.968) and KIRC (0.993 vs.\ 0.987). These were the three types with the lowest DE-ranking AUC; the differences were small and rest on five folds. In the other six, the two selectors were within 0.005 at $K = 10$, and the rule kept DE. The genes of the six DE panels were reproducible: the top five genes of each CV training fold overlapped the final panel with mean Jaccard indices of 0.57--0.80. The three hybrid panels were not (0.14--0.40; Table~\ref{tab:stage}): different five-gene sets reached similar AUC, so for these types the individual genes are interchangeable and should not be read as specific markers. In the held-out patients, AUC was 0.993--1.000 in all seven types that could be tested, sensitivity was 0.88--1.00 and specificity was 1.0 throughout. The held-out sets were small, however (5--19 normals), so the lower limits of the specificity CIs are 0.48--0.82 and specificity is estimated imprecisely. The panels also detected development stage~II and stage~III--IV tumours, which had never been used for fitting (sensitivity 0.92--1.00; Figure~\ref{fig:stage}b), except THCA stage~II (0.78). They are therefore tumour versus adjacent-normal signatures that already work at stage~I, not markers specific to early-stage disease.

\begin{table}[!htbp]
\caption{Five-gene stage~I tumour vs.\ adjacent-normal panels. Dev., development patients (stage~I / normal); AUC, ROC AUC; held-out, patients from the locked test set of the cancer-type task (sensitivity and specificity as counts); Stab., mean Jaccard index between the top five genes of each CV training fold and the final panel; --, fewer than five held-out samples of one kind. Per-gene DE statistics and fold selection counts are in Table~S9.\label{tab:stage}}
\small
\begin{tabularx}{\linewidth}{lllLccccc}
\toprule
\textbf{Type} & \textbf{Dev.} & \textbf{Selector} & \textbf{Panel (rank order)} & \textbf{Stab.} & \textbf{CV AUC} & \textbf{Held-out AUC} & \textbf{Sens.} & \textbf{Spec.}\\
\midrule
BRCA & 148/94 & DE & COL10A1, MMP11, COL11A1, HSD17B6, LINC01614 & 0.73 & 0.996 & 0.994 & 23/26 & 19/19\\
COAD & 49/33 & DE & FOXQ1, ETV4, KRT80, CDH3, SIM2 & 0.69 & 1.000 & 1.000 & 19/19 & 8/8\\
KIRC & 183/61 & hybrid & FAM245A, PIK3C2G, GPC5, LINC01983, AC104237.2 & 0.14 & 0.993 & 1.000 & 50/50 & 11/11\\
KIRP & 123/24 & DE & AQP6, UMOD, DDN, HEPACAM2, CRHBP & 0.69 & 1.000 & 1.000 & 29/29 & 8/8\\
LIHC & 107/39 & DE & PLVAP, COL15A1, GABRD, CDKN3, THBS4 & 0.57 & 0.997 & 1.000 & 26/26 & 11/11\\
LUAD & 178/49 & DE & PYCR1, FAM83A, ETV4, STX11, OTUD1 & 0.80 & 1.000 & 1.000 & 41/42 & 10/10\\
LUSC & 169/48 & DE & FAM83B, TFAP2A, SLC2A1, KCNMB2-AS1, KIF23 & 0.57 & 1.000 & -- & -- & --\\
STAD & 48/32 & hybrid & MTHFD1L, NFE2L3, NUDT5, B3GNTL1, KNOP1 & 0.32 & 0.993 & -- & -- & --\\
THCA & 209/54 & hybrid & LRATD1, ENTPD1, CDH6, SERINC2, GABRB2 & 0.40 & 0.985 & 0.993 & 51/54 & 5/5\\
\bottomrule
\end{tabularx}
\end{table}

\section{Discussion}

Our four aims addressed the four questions raised in the Introduction about classifiers and gene signatures built from TCGA RNA-seq data with the quality-aware hybrid selector: (i)~whether ensembles improve on a single well-regularised learner, (ii)~how much the accuracy measured with random splits depends on the tissue source site, (iii)~whether a panel selected in TCGA keeps its accuracy in an independent cohort, and (iv)~whether small gene panels can separate stage~I tumours from adjacent normal tissue when they are tested without selection leakage. The results bear mainly on how TCGA-derived classifiers should be evaluated, in particular how panel size, site-grouped validation and external transfer change what can be concluded; the biological results, namely the shared tumour programme and the five-gene panels, are candidate findings that need independent validation. We discuss the four aims in turn and then the limitations of the study.

\subsection{The Value of an Ensemble Depends on Panel Size}

With the 1000-gene panel, none of the other four learners (LightGBM, random forest, linear SVM and MLP) and none of the three ensembles (soft voting, stacking and the hierarchical model) reliably improved on L2-regularised logistic regression, which was also the best-calibrated model (Table~\ref{tab:models}, Figure~\ref{fig:models}). Stacking came closest. With gene selection nested inside the inner folds, it gained 0.0046 in balanced accuracy in CV and was better than LR in four of the five folds, but the gain stayed below the margin that we had fixed before the analysis (Figure~\ref{fig:models}c). On the test set it gained 0.006, which was not significant. With the minimal 100-gene panel, the picture changed: stacking gained 0.007 on average and was better in every fold. Ensembles therefore appear to help mainly when the panel is small. A plausible explanation is that, with few genes, each learner captures a different part of a limited signal and makes different errors, which the meta-learner can combine; with 1000 well-chosen genes, a single linear model already captures most of the signal, and little is left for the ensemble to add.

Gains of this size must also be weighed against their cost. Stacking requires all five learners, 25 additional inner fits per outer fold and, when selection is nested, a selector run in every inner training set, and it returned less than one percentage point of balanced accuracy. Among single learners the cost argument is even clearer: LightGBM took more than 100 times longer to fit than LR (17.6 vs.\ 0.15~min per outer fold) and was less accurate. These comparisons concern fixed, untuned settings. Tuning might improve LightGBM or the MLP, so we do not conclude that LR is inherently superior, only that none of the configurations we tested outperformed it. This agrees with the long-standing observation that linear models perform well on high-dimensional expression data once the features are well chosen~\cite{saeys2007,hastie2009}: expression data contain many correlated genes, and a strongly regularised linear model is less prone to overfit them than a flexible one.

Calibration is the second reason to prefer LR. A model is well calibrated when its predicted probabilities match observed frequencies: among tumours assigned a probability of about 0.9, about 90\% should be classified correctly. This matters if probabilities are used to flag uncertain cases for further review~\cite{niculescu2005}, and current reporting guidance for clinical prediction models asks for calibration to be reported alongside discrimination~\cite{collins2024}. LR had the lowest calibration error in CV and on the test set, whereas the random forest and the linear SVM were clearly worse calibrated (Table~\ref{tab:models}, Figure~\ref{fig:models}d).

Our test-set balanced accuracy of 0.943 is within the range of 90--96\% accuracy reported for TCGA classifiers~\cite{li2017,mostavi2020} and close to the median cross-validated accuracy (95.5\%) of 20 RNA-seq tissue-of-origin classifiers reviewed recently~\cite{stancl2023}. Those studies used other cohorts, classes and random splits and mostly reported plain accuracy, which is usually higher than balanced accuracy when classes are imbalanced, so the figures are not directly comparable. Stacked deep ensembles have reached about 99\% accuracy, but for only five cancer types arising in different organs~\cite{mohammed2021}, an easier task. With 33 types and a strong linear baseline, stacking added at most 0.007 in our hands.

\subsection{Remaining Errors and Site Effects Point to the Same Pair}

The hierarchical model first assigned each tumour to a group of similar cancer types and then to a type within that group. It did not help (Figure~\ref{fig:sites}a), because the flat model's remaining errors were concentrated in pairs that are molecularly close, and a specialist model trained within the group faces the same ambiguity with the same genes. The two pairs with the most confusion, colon vs.\ rectal and stomach vs.\ oesophageal adenocarcinoma, are known to be biologically similar. TCGA's colorectal analysis found colon and rectal adenocarcinomas indistinguishable at the molecular level, apart from their anatomical location~\cite{tcga2012crc}. Oesophageal adenocarcinomas closely resemble the chromosomally unstable subtype of gastric adenocarcinoma~\cite{tcga2017esca}, and the two cluster together in pan-cancer analyses~\cite{hoadley2018}. Part of the remaining error may therefore be a limit set by biology rather than a weakness of the classifier.

Site-grouped validation adds a second explanation for the colorectal pair. In TCGA, each tumour carries a tissue source site (TSS) code that identifies the collection that contributed it, and every TSS code contributed tumours of a single cancer type. With random splits, tumours from the same collection appear in both the training and the validation folds. A classifier can then partly recognise the collection rather than the tumour type, and still be scored as correct. The signals that identify a collection may be technical, such as sample handling, processing and sequencing batch, or biological, such as differences in the patients each collection contributed. Site grouping keeps every collection entirely in one fold, so it shows how much accuracy depends on such signals, but it cannot tell technical from biological sources apart.

For COAD and READ, whose expression profiles are otherwise nearly identical, the collection may be one of the few remaining clues, so site-associated signals matter most for this pair. When collections were held out, rectal recall halved (from 0.68 to 0.33) and colon recall fell from 0.85 to 0.58 (Figure~\ref{fig:sites}b; Table~S3, Figure~S2). Rectal recall fell under every assignment of sites to folds that we tried (Table~S4). Part of the rectal drop may reflect how the rectal tumours were distributed across collections: one collection contributed 44\% of the rectal development tumours, so the site-grouped fold that held it out also removed almost half of the rectal training data. Concentration alone did not cause large drops, however: one collection contributed 90\% of ACC tumours, yet ACC recall did not change (Table~S3). Outside colorectal cancer, the classifier was robust to unseen sites: without COAD, READ and LAML, balanced accuracy fell by only 0.003--0.008 across the five assignments (Figure~\ref{fig:sites}c, Table~S4). Holding out sites also affected gene selection: panels selected in site-grouped training folds agreed less with each other than panels selected in random-split folds (mean Jaccard index 0.64--0.72 vs.\ 0.74--0.83; Table~S5, Figure~S1).

These findings matter for how TCGA classifiers are evaluated. Most published studies, including those in Table~\ref{tab:related}, estimated accuracy with random splits~\cite{li2017,mostavi2020,mohammed2021}, which place tumours from the same collection in both the training and the validation data. Our results show that such splits overstate accuracy, mainly where site and label are entangled, as for colon and rectal cancer. A similar problem has been reported in digital histology, where site-specific signatures in TCGA slides inflated the accuracy of deep learning models unless sites were kept apart during validation~\cite{howard2021}. Site-grouped validation, with the selector rerun inside each fold, needs only one additional cross-validation of the whole pipeline and should therefore become standard practice.

\subsection{External Transfer}

The breast-cancer subtype panel kept high accuracy on SCAN-B, even though the two cohorts differ in sequencing platforms, expression units (TPM vs.\ FPKM) and patient selection (a multi-centre research collection vs.\ a population-based cohort). Its balanced accuracy of 0.855 came within 0.03 of the 50 PAM50 genes (Figure~\ref{fig:external}a), which are an upper reference because the SCAN-B subtypes are themselves PAM50 calls. LR transferred markedly better than LightGBM, for the selected panels and for the PAM50 genes alike. One possible reason is that tree models split on fixed expression thresholds learned in the training cohort, and these thresholds move when the expression distribution changes between cohorts, whereas a linear model depends on the overall direction of change across many genes. By comparison, tissue-of-origin classifiers trained on TCGA lost several percentage points of accuracy in independent cohorts~\cite{zhao2020,divate2022}. SCAN-B has also been used to train RNA-seq single-sample predictors of PAM50 subtype, which agreed well with a clinical assay in external cohorts~\cite{staaf2022}.

The quality-aware panel performed comparably to the baseline panel on SCAN-B, not significantly better, and neither panel lost accuracy on lower-quality samples (Table~S6). This is consistent with our previous study~\cite{paper1}, where we claimed that the quality-aware selector picks fewer quality-associated genes at no cost in accuracy, not that it improves robustness in other cohorts.

One limitation of the external test became clear. The lncRNAs and other non-coding genes that Boruta selected in TCGA were largely absent from SCAN-B's gene annotation: of the six lncRNAs among the top 60 genes of the quality-aware ranking, five were missing, compared with three of 51 protein-coding genes (Figure~\ref{fig:external}b, Table~S6). Such genes are widely and often cancer-specifically expressed~\cite{iyer2015,yan2015,han2014,kalyanasundaram2012}, which is why the selector can pick them, but here they were also the genes most often missing from the other cohort's annotation. Whether a gene is quantified at all depends on the annotation used by each processing pipeline, so a non-coding gene measured in TCGA may simply not exist in another cohort's data. As a result, the panels used on SCAN-B were almost entirely protein-coding, and the external test validates the protein-coding core of the panels, not their non-coding contribution. Non-coding genes were also the least stable part of the panels across resamples (Table~S5). Panels intended for use in other cohorts will therefore need either annotation-matched processing pipelines or gene selection restricted to genes measured on the target platform.

Finally, all models, including those built on the PAM50 genes, were more accurate on the NextSeq than on the HiSeq tumours of SCAN-B, although the two groups had nearly identical subtype mix, detected-gene counts and clinical characteristics (Figure~\ref{fig:external}c, Table~S6). Because the gap appeared with all 11 models, it most likely reflects a platform-level difference within SCAN-B rather than a property of the selected panels. This platform effect was also larger than the effect of the choice of panel: for LR with the quality-aware panel, accuracy differed by 0.065 between platforms, compared with 0.007 between the quality-aware and the baseline panel. Only 304 tumours were sequenced on NextSeq, however, so the size of the platform gap is uncertain (95\% CI 0.022--0.106).

\subsection{Tumour Versus Adjacent-Normal Panels}

Many tumour versus normal signatures have been reported from TCGA, so the value of our panels lies less in the individual genes than in how they were tested. A common weakness of such signatures is selection leakage: genes are chosen using all samples, and the same samples are then used to measure accuracy, which inflates the result. Here, discovery was confined to development patients, and gene ranking was repeated inside every CV fold, so the validation patients of each fold never influenced which genes were chosen. The panels were then scored once in held-out patients and in later-stage tumours that had never been used for fitting. Under these conditions, five genes still sufficed in every eligible cancer type (Table~\ref{tab:stage}, Figure~\ref{fig:stage}).

Stage~I tumours differed from adjacent normal tissue in a way that was largely shared across types. Genes for extracellular-matrix remodelling and cell proliferation were up-regulated, and genes for the specialised functions of each normal tissue, such as secretion and signalling, were down-regulated; 1911 genes changed in the same direction in at least five of the nine types (Figure~\ref{fig:de}; Tables~S7 and S8). Even at stage~I, therefore, tumours already show both the gain of a common tumour programme and the loss of normal tissue identity.

The comparison of selectors carries a lesson about how gene lists should be read. In the three types where the task was hardest (STAD, THCA and KIRC), the hybrid selector gave a slightly higher AUC than ranking by differential expression, but its genes changed from fold to fold (mean Jaccard index 0.14--0.40), whereas the DE panels were reproducible (0.57--0.80; Table~\ref{tab:stage}, Figure~\ref{fig:stage}a). In these types, many genes separate tumour from normal tissue about equally well, for example because they are co-regulated, and the five genes chosen are only one of several interchangeable sets. For these panels, the accuracy is the robust result, not the gene list, and the individual genes should not be read as specific markers.

Several genes of the reproducible panels have established roles in their tumour type. COL11A1 marks activated cancer-associated fibroblasts across cancers~\cite{jia2016}. FOXQ1 is overexpressed in colorectal cancer~\cite{kaneda2010}. PLVAP marks the onco-fetal endothelium of hepatocellular carcinoma~\cite{sharma2020}. In KIRP, UMOD and AQP6 are genes of the normal kidney tubule that are lost in the tumour ($\log_2$ fold change $-13.0$ and $-8.7$; Table~S9); they detect the tumour through the absence of normal tissue rather than through a tumour-specific gain. Because the panels detected stage~II and stage~III--IV tumours almost as well as stage~I tumours, except THCA stage~II (Figure~\ref{fig:stage}b), they mark tumour tissue from stage~I onward rather than early-stage disease specifically.

Two caveats limit the clinical reading of these results. First, adjacent normal tissue is not healthy tissue: it is taken from the same organ as the tumour and carries its own tumour-induced changes~\cite{aran2017}, so the panels may perform differently against tissue from people without cancer. Second, the held-out sets contained only 5--19 normal samples per type, so specificity is estimated imprecisely, with lower confidence limits between 0.48 and 0.82 (Figure~\ref{fig:stage}c). The panels are therefore candidate tissue signatures that discriminate tumour from adjacent-normal tissue in TCGA. They have not been validated against healthy-control tissue or in independent cohorts, and they are not screening or diagnostic tests.

\subsection{Limitations}

The study has limitations in its data, its evaluation design and its scope. All analyses except the external transfer used TCGA, so the findings may not hold for cohorts collected or processed differently. The locked test set was used for the second time, as declared in advance. Its first use, in our previous study, involved the same LR model and panel, and no design decision in the present study depended on test results; nevertheless, the test set is no longer an untouched estimate. In addition, the test-set stacking result used the non-nested meta-learner. All learners had fixed, untuned settings, so the model comparison concerns these configurations rather than the algorithms in general.

Site grouping establishes that random splits are optimistic in a site-associated way, but it cannot separate technical from biological sources, and cancer types collected at a single site, such as LAML, cannot be tested. Several analyses were added after the main results and are reported as exploratory: the site analysis without COAD, READ and LAML, the further site-to-fold assignments, the selection-stability analysis and the SCAN-B platform comparison. In the external test, SCAN-B provides PAM50 calls rather than independent histological subtypes, and genes were matched by symbol across annotation versions; most non-coding genes of the TCGA panels were absent from SCAN-B's annotation.

For the five-gene panels, the comparison was with adjacent-normal rather than healthy tissue, and the held-out sets contained few normal samples. The genes of the hybrid panels were not stable across folds, two eligible types (LUSC and STAD) had too few held-out samples to be tested, and no panel has been validated in an independent cohort. Validation in cohorts with healthy-control tissue and with matched gene annotation would address the main open questions.

\section{Conclusions}

With 1000 genes from a quality-aware hybrid Boruta selector, L2-regularised logistic regression was the most accurate single learner, and no ensemble improved on it by the pre-declared margin. It was also better calibrated and far cheaper than the ensembles. With gene selection nested in every training fold, stacking helped consistently only with a minimal 100-gene panel. Site-grouped validation, with the selector rerun inside each fold and repeated over five site-to-fold assignments, showed that the classifier is robust to unseen sites except in colorectal cancer, where random splits overstate accuracy; whether the underlying site signal is technical or biological cannot be decided from TCGA alone. The breast-cancer subtype panel transferred to an independent population-based cohort with accuracy comparable to the baseline panel, although most of its non-coding genes were absent from the external data. Five-gene panels, derived without access to held-out patients, separated tumours from adjacent normal tissue from stage~I onward; their accuracy was robust, but for three types their genes were interchangeable. We recommend site-grouped and external validation as routine checks for TCGA-derived classifiers. Future work should test the stage~I panels in independent cohorts with healthy-control tissue, and should restrict selection to genes measured by the target platform.

\section*{Supplementary Materials}
Supplementary Methods S1 (a self-contained description of the quality-aware hybrid selector, with Supplementary Figures S1--S3) is available as an ancillary file of this preprint. Supplementary Tables S1--S9 are archived with the code (Data Availability Statement): Table S1, model comparison, CV metrics per fold, test-set results and decision rules; Table S2, nested and non-nested stacking; Table S3, site-grouped CV, samples and tissue source sites per cancer type, folds, pooled drop and per-class recall; Table S4, site-grouped CV under five site-to-fold assignments; Table S5, selection stability and per-gene selection frequency; Table S6, SCAN-B results by panel, classifier, platform and subtype, confusion matrices and unmatched genes; Table S7, DE results per cancer type; Table S8, significant GO and KEGG terms; Table S9, five-gene panels with DE statistics, fold selection counts and CV, later-stage and held-out results. Figure S1, selection stability against panel size; Figure S2, per-class recall under random-split and site-grouped CV; Figure S3, SCAN-B confusion matrices.

\section*{Author Contributions}
Tareque Mohmud Chowdhury conceived and designed the approach, performed the experiments, analysed the data and wrote the manuscript. Abu Raihan Mostofa Kamal supervised the study and revised the manuscript. Both authors have read and approved the final version of the manuscript.

\section*{Funding}
This research received no external funding.

\section*{Code and Data Availability Statement}
TCGA RNA-seq and clinical data are available from the NCI Genomic Data Commons (\url{https://portal.gdc.cancer.gov}). SCAN-B data are available from GEO under accession GSE96058. The code is available from the corresponding author upon reasonable request.

\section*{Acknowledgments}
The results shown here are in part based upon data generated by the TCGA Research Network (\url{https://www.cancer.gov/tcga}) and by the SCAN-B initiative. The research questions, methodology, experimental design, result analyses, and conclusions are the authors’ own. The AI assistant Claude (Anthropic) was used to check and refine the language of the author written manuscript, which was reviewed and verified by the authors, who take full responsibility for the accuracy and integrity of the research article.

\section*{Conflicts of Interest}
The authors declare no conflicts of interest.

\section*{Abbreviations}
\noindent
\begin{tabular}{@{}ll}
AUC & Area under the receiver operating characteristic curve\\
CV & Cross-validation\\
DE & Differential expression\\
ECE & Expected calibration error\\
LR & Logistic regression\\
QC & Quality control\\
SCAN-B & Sweden Cancerome Analysis Network--Breast\\
TCGA & The Cancer Genome Atlas\\
TPM & Transcripts per million\\
TSS & Tissue source site
\end{tabular}

\bibliographystyle{IEEEtran}
\bibliography{refs}
\end{document}